\documentclass[10pt,twocolumn,letterpaper]{article}

\usepackage{cvpr}              % To produce the CAMERA-READY version

\usepackage{graphicx}
\usepackage{amsmath}
\usepackage{amssymb}
\usepackage{booktabs}
\usepackage{colortbl}  
\usepackage{xcolor}
\usepackage{array}

\usepackage{makecell}

\definecolor{hollywoodcerise}{rgb}{0.96, 0.0, 0.63}
\definecolor{lasallegreen}{rgb}{0.03, 0.47, 0.19}
\definecolor{hanpurple}{rgb}{0.32, 0.09, 0.98}
\definecolor{green(pigment)}{rgb}{0.0, 0.65, 0.31}

\usepackage{hyperref}
\hypersetup{colorlinks,linkcolor={red},citecolor={hollywoodcerise},urlcolor={red}}  

\usepackage[capitalize]{cleveref}
\crefname{section}{Sec.}{Secs.}
\Crefname{section}{Section}{Sections}
\Crefname{table}{Table}{Tables}
\crefname{table}{Tab.}{Tabs.}

\def\cvprPaperID{3247} % Enter the CVPR Paper ID here
\def\confName{CVPR}
\def\confYear{2023}

\begin{document}

\title{
Learning Spatial-Temporal Implicit Neural Representations for Event-Guided Video Super-Resolution}

\author{Yunfan~Lu$^{1}$~$^{*}$  \quad Zipeng~Wang$^{1}$ \thanks{These authors are co-first authors.}
\quad Minjie~Liu$^{1}$~$^{\dag}$ \quad Hongjian~Wang$^{2}$ \thanks{These authors are co-second authors.}
\quad Lin~Wang$^{1,3}$\thanks{Corresponding author}\\
$^{1}$AI Thrust, HKUST(GZ) \quad  $^{2}$Shenzhen International Graduate School, Tsinghua University \\
\quad $^{3}$Dept. of Computer Science and Engineering, HKUST\\
{\tt\small \{ylu066,zwang253,mliu942\}@connect.hkust-gz.edu.cn, whj20@mails.tsinghua.edu.cn, linwang@ust.hk}
}

\maketitle

\begin{abstract}
Event cameras sense the intensity changes asynchronously and produce event streams with high dynamic range and low latency. 
This has inspired research endeavors utilizing events to guide the challenging video super-resolution (VSR) task.
In this paper, we make the \textbf{first} attempt to address a novel problem of achieving VSR at random scales by taking advantages of the high temporal resolution property of events. This is hampered by the difficulties of representing the spatial-temporal information of events when guiding VSR.
To this end, we propose a novel framework that incorporates the spatial-temporal interpolation of events to VSR in a unified framework. 
Our key idea is to learn implicit neural representations from queried spatial-temporal coordinates and features from both RGB frames and events. 
Our method contains three parts. Specifically, the Spatial-Temporal Fusion (\textbf{STF}) module first learns the 3D features from events and RGB frames. Then, the Temporal Filter (\textbf{TF}) module unlocks more explicit motion information from the events near the queried timestamp and generates the 2D features. Lastly, the Spatial-Temporal Implicit Representation (\textbf{STIR}) module recovers the SR frame in arbitrary resolutions from the outputs of these two modules.
In addition, we collect a real-world dataset with spatially aligned events and RGB frames. Extensive experiments show that our method significantly surpasses the prior-arts and achieves VSR with random scales, \eg, 6.5.
Code and dataset are available at \url{https://vlis2022.github.io/cvpr23/egvsr}.

\end{abstract}

\begin{figure}
    \centering
    \includegraphics[width=\columnwidth]{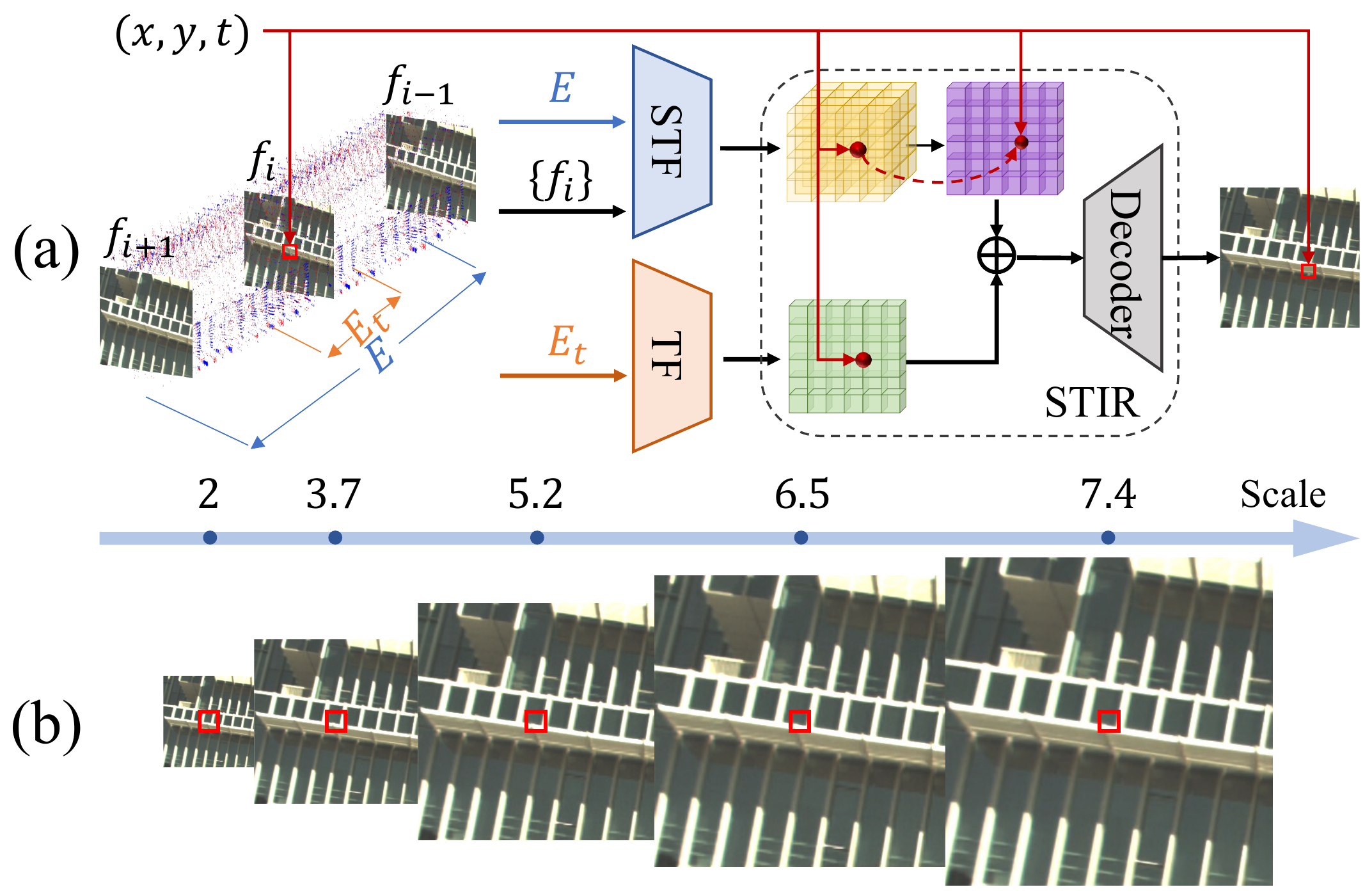}
    \vspace{-18pt}
    \caption{(a) Our method learns implicit neural representations (INR) from the queried spatial-temporal coordinates (STF) and temporal features (TF) from RGB frames and events. 
    % incorporates the spatial-temporal interpolation of events to learn implicit representations for VSR. 
    (b) An example of VSR with random scale factors, \eg, 6.5, by our method.}
     \vspace{-15pt}
    \label{fig:0-CoverFigure-V3}
\end{figure}
\vspace{-17pt}
\section{Introduction}
\label{sec:intro}

Video super-resolution (VSR) is a task of recovering high-resolution (HR) frames from successive multiple low-resolution (LR) frames. 
Unlike LR videos, HR videos contain more visual information, \eg, edge and texture, which can be very helpful for many tasks, \eg, metaverse~\cite{zhou2022vetaverse}, surveillance~\cite{ijjina2019computer} and entertainment~\cite{canedo2019facial}. 
However, VSR is a highly ill-posed problem owing to the loss of both spatial and temporal information, especially in the real-world scenes~\cite{wang2020deep,lu2022all,chan2021basicvsr++, chan2021basicvsr}.
Recently, deep learning-based algorithms have been successfully applied to learn the intra-frame correlation and temporal consistency from the LR frames to recover HR frames, \eg, 
DUF\cite{jo2018deep}, EDVR\cite{wang2020basicsr}, RBPN \cite{Haris2019rbpn}, BasicVSR \cite{chan2021basicvsr}, BasicVSR++ \cite{chan2021basicvsr++}. 
However, due to the lack of inter-frame information, these methods are hampered by the limitations of modeling the spatial and temporal dependencies and may fail to recover HR frames in complex scenes. 

Event cameras are bio-inspired sensors that can asynchronously detect the per-pixel intensity changes and generate event streams with low latency (1$us$) and high dynamic range (HDR) compared with the conventional frame-based cameras (140$dB$ vs. 60$dB$)~\cite{scheerlinck2019ced,zheng2023deep}.
This has sparked extensive research in reconstructing image/video from events~\cite{gallego2020event,wang2020eventsr,wang2021evdistill,mostafavi2021e2sri,wang2020event,zou2021learning}. 
However, the reconstructed results are less plausible due to the loss of visual details, \eg, structures, and textures.
As a result, a recent work has utilized events for guiding VSR~\cite{jing2021turning}, trying to `inject' energy from the event-based to the frame-based cameras. 
It leverages the high-frequency event data to synthesize neighboring frames, so as to find correspondences between consecutive frames.
However, it only treats video frames in discrete ways with 2D arrays of pixels and up-samples them at a fixed up-scaling factor~\eg, $\times 2$ or $\times 4$.
This causes inconvenience and inflexibility in the applications of SR videos, which often require arbitrary resolutions, \ie, random scales. 

Recently, some works tried to learn continuous image representations with arbitrary resolutions, \eg, LIIF\cite{chen2021liif}, taking 2D queried coordinates and 2D features as input to learn an implicit neural representation (INR).
VideoINR \cite{chen2022videoinr}, on the other hand, decodes the LR videos into arbitrary spatial resolutions and frame rates by learning from the spatial coordinates and temporal coordinates, respectively.
However, it is still unclear how to leverage events to guide learning spatial-temporal INRs for VSR. 
This is hampered by two challenges. Firstly, although event data can benefit VSR with its high-frequency temporal and spatial information, the large modality gap between the events and video frames makes it challenging to use INRs to represent the 3D spatial-temporal coordinates with event data.
Moreover, there lacks HR real-world dataset with spatially well-aligned events and frames. 

In this paper, we make the \textbf{first} attempt to address a novel problem of achieving VSR at random scales by taking advantage of the high-temporal resolution property of events.
Accordingly, we propose a novel framework that subtly incorporates the spatial-temporal interpolation from events to VSR in a unified framework, as shown in Fig.~\ref{fig:0-CoverFigure-V3}. 
Our key idea is to learn INRs from the queried spatial-temporal coordinates and features from both the RGB frames and events.
Our framework mainly includes three parts.
The Spatial-Temporal Fusion (\textbf{STF}) branch learns the spatial-temporal information from events and RGB frames (Sec.~\ref{subsec:stf_branch}). The shallow feature fusion and deep feature fusion are employed to narrow the modality gap and fuse the events and RGB frames into 3D global spatial-temporal representations. 
Then, the Temporal Filter (\textbf{TF}) branch further unlocks more explicit motion information from events. It learns the 2D event features from events nearing the queried timestamp~(Sec.~\ref{sec:temporal_filter_branch}). 
With the features from the STF and TF branches, the Spatial-Temporal Implicit Representation (\textbf{STIR}) module decodes the features and recovers SR frames with arbitrary spatial resolutions(Sec.~\ref{subsec:stir_module}). That is, given the arbitrary queried coordinates, we apply 3D sampling and 2D sampling to the fused 3D features and event data separately. Finally, the sampling features are added and fed into a decoder, and generate targeted SR frames.
In addition, we collect a real-world dataset with a spatial resolution of $3264\times 2248$, in which the events and RGB frames are spatially aligned.
Extensive experiments on two real-world datasets show that our method surpasses the existing methods by \textbf{1.3 dB}.

In summary, the main contributions of this paper are fivefold: 
(\textbf{I}) Our work serves as the \textbf{first} attempt to address a non-trivial problem of learning INRs from events and RGB frames for VSR at random scales. 
(\textbf{II}) We propose the STF branch and the TF branch to model the spatial and temporal dependencies from events and RGB frames. 
(\textbf{III}) We propose the STIR module to reconstruct RGB frames with arbitrary spatial resolutions. 
(\textbf{IV}) We collect a high-quality real-world dataset with spatially aligned events and RGB frames. 
(\textbf{V}) Our method significantly surpasses the existing methods and achieves SR with random-scales, \eg, 6.5.

\begin{figure*}[t!]
\centering
\includegraphics[width=1\linewidth]{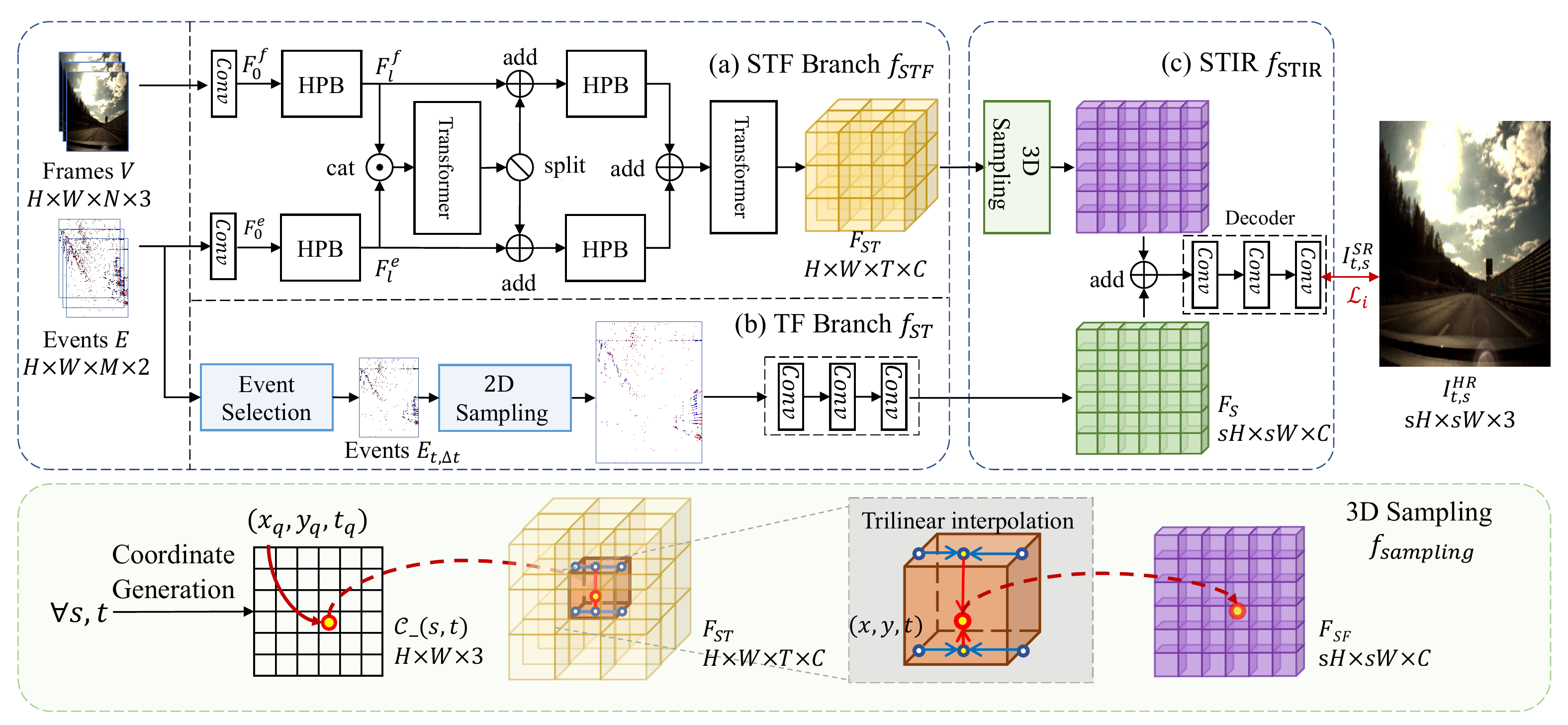}
\vspace{-18pt}
\caption{Overview of our proposed framework. 
Our method consists of three parts, the Spatial-Temporal Fusion (STF), the Temporal Filter (TF) and Spatial-Temporal Implicit Representation (STIR).
These three parts are shown in (a) (b) (c) of this figure, respectively.
Details of STF, TF and STIR are described in this Sec.\ref{subsec:stf_branch}, Sec.\ref{sec:temporal_filter_branch} and Sec.\ref{subsec:stir_module}, respectively.
}
\label{fig:0-Overall-Framework}
\centering
\vspace{-10pt}
\end{figure*}

\section{Related Work}
\label{sec:related_work}

\noindent{\bf Event-guided Video and Image SR} 
Recently, event data has shown the potential to guide the image or video SR. 
eSL-Net\cite{wang2020esl} and EvIntSR\cite{han2021evintsr} focus on employing events to guide the single image SR. Specifically, eSL-Net\cite{wang2020esl} feeds both the events and LR image to a sparse learning framework to recover an SR image. EvIntSR\cite{han2021evintsr} first reconstructs the latent frames from the events and LR image, which are then fused to reconstruct the SR image.
Differently, event-guided VSR takes consecutive frames and events as inputs and models both the spatial and temporal information. A recent work~\cite{jing2021turning} proposed a two-stage method by 1) utilizing events to interpolate the LR video to get a high-frequency video and 2) rebuilding HR key frames. However, it encodes video frames in discrete ways and only up-samples videos at a fixed upscale factor, \eg, $\times 2$. We make the first attempt to achieve VSR at random scales by learning the spatial-temporal implicit representations from events and video frames.

\noindent{\bf Video Super-Resolution (VSR)}
The dominant research for VSR mainly focuses on designing learning pipelines~\cite{liu2022videosr}, concerning the feature learning, frame alignment and multi-frame fusion~\cite{caballero2017real,liu2018learning,yang2021real,yue2022real,nah2019ntire,wang2021unsupervised,tian2020tdan,li2019fast,isobe2020video}.
For example, Bao \textit{et.al} \cite{Bao2021memc} employed the motion compensation to achieve the frame alignment, while EDVR\cite{wang2020basicsr} proposes the deformable convolutions after extracting the features from the input frames. To address the feature propagation and alignment problem effectively, BasicVSR\cite{chan2021basicvsr} and BasicVSR++\cite{chan2021basicvsr++} proposed a succinct pipeline based on the bidirectional propagation and optical flow. To better exploit the temporal information of video frames, RBPN \cite{Haris2019rbpn} extracts and propagates the spatial and temporal information of consecutive frames in a recurrent back-projection manner. Inspired by the random image upsampling,  VideoINR\cite{chen2022vinr} represents the frames with implicit representations and thus makes it possible to learn random-scale VSR. However, these works focus on a single modality. Differently, in our work, we make the first attempt to address a novel problem of achieving VSR at random scales by taking advantage of the high-temporal resolution property of events.

\noindent{\bf Implicit Neural Representation (INR)} 
It also called the coordinate-based representation, aims at parameterizing signals, \eg, images and audio, in a continuous way via neural networks~\cite{sitzmann2020siren}. INR has been widely applied to 3D scene representation\cite{park2019deepsdf, mildenhall2021nerf} and generative models\cite{niemeyer2021giraffe, jain2022dreamfields}, etc. 
Recently, INRs have been extensively studied for image and video SR. For instance,  LIIF\cite{chen2021liif} achieves image SR with random scales given the 2D queried coordinates and 2D features.
JIIF\cite{tang2021jiif} uses HR images to guide the interpolation weights and values of LR depth maps. 
VideoINR \cite{chen2022videoinr} decodes the LR and low-frame-rate videos into an arbitrary spatial resolution and frame rate with three INRs. It adopts two INR functions to learn the spatial coordinates and the temporal coordinates, respectively. These two INR functions are then used to generate a motion flow field, which is applied back to warp the encoded features. Then, the warped features are decoded to recover the SR frame by the spatial INR function.
Differently, considering the large modality gap between the events and video frames, we accordingly propose a novel INR module to directly represent the 3D spatial-temporal coordinates from event data.

\vspace{-6pt}
\section{The Proposed Framework}
\label{sec:methods}
\noindent \textbf{Event Representation}
\label{subsec:event_representation} As event streams are sparse points, we first describe how to stack them into the fixed-size representations as inputs of our framework. 
Events are produced by detecting variations in the log intensity of each pixel.
An event $e=(x,y,t,p)$ is triggered and recorded when the logarithmic brightness change exceeds a certain threshold $\theta$ at pixel $(x,y)$.
This process can be described as Eq.\ref{eq:event_generation}, where $\Delta L = log(I^l_t + n) - log(I^l_{t-\Delta t} + n)$, $n$ is noise, $I^l_t$ and $I^l_{t-\Delta t}$ are intensity values in linear domain at timestamps $t$ and $t-\Delta t$, respectively.

\begin{equation}
    \begin{aligned}
        p = \left\{
            \begin{aligned}
                +1, &\Delta L > \theta \\
                -1, &\Delta L < -\theta
            \end{aligned}
        \right.
    \end{aligned}
    \label{eq:event_generation}
\end{equation}

According to Eq.\ref{eq:event_generation}, the relation between frames $I^l_{t_0}(x,y)$ and $I^l_{t_1}(x,y)$ at timestamps $t_0$ and $t_1$ can be formulated as Eq.\ref{eq:frame_relation_with_event}, where $p$ is the polarity of event at pixel $(x,y)$.

{\setlength\abovedisplayskip{-4pt}
\setlength\belowdisplayskip{3pt}
\begin{equation}
   \begin{aligned}
    % \vspace{-5pt}
        I^l_{t_1}(x,y) = I^l_{t_0}(x,y)\times \exp({\theta \int_{t_0}^{t_1} p~ dt})
    \vspace{-5pt}
    \end{aligned}
    \label{eq:frame_relation_with_event}
\end{equation}}

Events record the intensity changes with higher temporal resolution than frames, which is advantageous for VSR task~\cite{jing2021turning}.
We split events into $M$ moment segments~\cite{wang2020event} with a shape of $H\times W\times M\times 2$ as the input to our framework.
Each segment keeps events that take place within time window. The window size is set to be a small constant to preserve the temporal information of events.

\subsection{Overview}
\label{subsec:overall}
The overall framework of our method is depicted in Fig.\ref{fig:0-Overall-Framework}, which consists of three parts: 
(I) spatial-temporal fusion (STF) branch,
(II) temporal filter (TF) branch, and 
(III) spatial-temporal implicit representation (STIR) module.
The input of our framework are spatially aligned RGB video frames $V=\{V_{0}...V_{i}...V_{n}\}$ and events $E \in R^{H\times W\times M\times 2}$, where 
$H \times W$ denotes the frame size.   
The output of our framework is a super-resolved video frame $I_{s,t}^{SR}$ with the up-sampling scale $s$ and the timestamp $t$. Note that the values of $s$ and $t$ can be freely adjusted. 
In practice, $s$ is a real number greater than 1, and $t$ can take the timestamps of all frames. 
The spatial resolution of the SR frame $I_{s,t}^{SR}$ is $sW \times sH$. 

Our framework consists of three major components.
The STF branch learns the holistic spatial-temporal information from events and RGB frames (Sec.\ref{subsec:stf_branch}).
Then, the TF branch unlocks more explicit motion information from events. 
It learns 2D event features from events nearing the queried timestamp $t$  (Sec.\ref{sec:temporal_filter_branch}).
Lastly, the STIR module decodes the features and recovers SR frames with arbitrary spatial resolutions (Sec.\ref{subsec:stir_module}).
We now describe these components in detail in the following sections.

\subsection{Spatial-Temporal Fusion (STF) Branch}
\label{subsec:stf_branch}

This module aims to extract spatial and temporal information from events $E$ and RGB frames $V$ to obtain a global spatial-temporal representation $F_{ST}$.
The representation is a 3D feature map of $H\times W\times T\times C$, where $T$ is the temporal dimension, and $C$ is the number of representation channels. The output $F_{ST}$ of STF branch $f_{STF}$ can be described as:
{\setlength\abovedisplayskip{6pt}
\setlength\belowdisplayskip{3pt}
\begin{equation}
    \begin{aligned}
        F_{ST} = f_{STF}(V, E)
    \end{aligned}
    \label{eq:stf}
\end{equation}}

Specifically, as depicted in Fig.~\ref{fig:0-Overall-Framework}(a), we first employ two $1 \times 1$ convolutional layers to obtain the initial frame feature map $F_{0}^f$ and the event feature map $F_{0}^e$ with same dimension. 
As stated in \cite{han2021evintsr}, shallow features preserve sharper details and local structural information while deeper features preserve more semantic information.
Thus, we design fusion blocks to aggregate both the shallow and deep features.

\noindent \textbf{Shallow feature fusion:}
As explored in \cite{lu2022transformer}, the residual architecture improves the model's capacity for representation and lessens the gradient vanishing issue.
For this reason, we first employ two high preserving blocks (HPB) as the basic feature extractors to extract two shallow feature maps $F_{l}^f$ and $F_{l}^e$ from the initial feature maps $F_{0}^f$ and $F_{0}^e$, respectively. 
After that, $F_{l}^f$ and $F_{l}^e$ are concatenated and then fed into a transformer-based fusion model to obtain a fused feature map so as to bridge the modality gap. The intuition behind this is that transformer has a larger perception field~\cite{vaswani2017attention,liu2022video,neimark2021video,arnab2021vivit,liu2022video}, which can be potentially used to model the global spatial and temporal dependencies from the frames and events. Then, the fused feature map is split into two parts in the channel dimension and added to $F_{l}^f$ and $F_{l}^e$ as the input for deep feature fusion.

\noindent \textbf{Deep feature fusion:}
After the shallow feature fusion, we again use two HPBs to extract deep features, which are then added and passed to the transformer-based fusion model to attain a 3D feature map $F_{ST}$.
Through shallow and deep feature fusion, we can better learn the temporal and spatial information from the events and frames.

\subsection{Temporal Filter (TF) Branch}
\label{sec:temporal_filter_branch}
Through the STF branch, the spatial-temporal feature information from both the frames and events can be effectively learned.
However, STF branch is insufficient to take full advantage of the high temporal resolution of event data. Therefore, we design the TF branch to explore more detailed motion information solely from events, which turns out to be effective in further enhancing the VSR performance, as demonstrated in our experiments (See Table~\ref{tab:ablation_TF_branch_shallow_feature_fusion}).

Intuitively, we design the TF branch $f_{TF}$  to capture the detailed motion information from events $E_{t,\Delta t}$ near the key frame at timestamp $t$, where $\Delta t$ is a small time interval. 
TF branch first selects events from $t - \Delta t$ to $t + \Delta t$ (Fig.~\ref{fig:0-Overall-Framework}(c)).
The selected events are interpolated and sent to three convolutional layers to learn the temporal features. 
Overall, the output $F_{T}$ of TF branch  $f_{TF}$ can be described as: 
{\setlength\abovedisplayskip{3pt}
\setlength\belowdisplayskip{3pt}
\begin{equation}
    \begin{aligned}
    F_{T} = f_{TF}(E_{t, ~\Delta t})
    \end{aligned}
    \label{eq:tf}
\end{equation}}

In Sec.\ref{ablation_discussion}, we show that STF branch can capture pixel intensities, while the TF branch can capture the motion details, \eg, edges and corners.

\subsection{Spatial-Temporal Implicit Representation}
\label{subsec:stir_module}

In this section, our goal is to learn continuous INRs for VSR based on the spatial-temporal feature map $F_{ST}$ and temporal feature map $F_{T}$. 
The INRs are then used to decode coordinates at time $t$ with scale factor $s$ into RGB values.
In this paper, we introduce the Spatial-Temporal Implicit Representation (STIR) module to accomplish the spatial-temporal VSR, as shown in Eq.\ref{eq:stir}.
We employ a simple-yet-effective 3D feature sampling and trilinear interpolation scheme to upsample $F_{ST}$ and $F_{T}$ to a desired resolution. A decoder, parameterized as a multi-layer CNN, is used to convert the interpolated features into RGB values. Fig.~\ref{fig:0-Overall-Framework}(c) depicts the detailed design of STIR.
The output SR frame $I^{SR}_{t,s}$ of the STIR $f_{STIR}$ can be formulated as: 
{\setlength\abovedisplayskip{1pt}
\setlength\belowdisplayskip{3pt}
\begin{equation}
    \begin{aligned}
    I^{SR}_{t,s} =f_{STIR}(F_{ST}, F_{T}), ~ \forall s, t 
    \end{aligned}
    \label{eq:stir}
\end{equation}}

\noindent \textbf{3D Feature Sampling:}
Here, we aim at generating a coordinate to make a query in grid form $F_{ST}$.
We uniformly sample a 3D coordinate grid, which can be expressed as $C_{t,s}$ with the dimension of $sH\times sW\times 3$. Formally, for any query $q$, the corresponding element $p_q$ in the 3D coordinate grid $C_{t,s}$ can be described as $p_q = (x_q, y_q, t_q)$, where $x_q \in [0, H], y_q \in [0, W], t_q \in [t_s, t_e]$, $t_s$ is the start time and $t_e$ is the end time of input.
For each coordinate $p_q = (x_q,y_q,t_q)$, we choose features of the nearest eight points around this coordinate in the 3D spatial-temporal feature $F_{ST}$ for interpolation.

\noindent \textbf{Feature Interpolation:}
Then, we compute the feature of a queried coordinate $p_q$ by using 3D interpolation techniques, such as trilinear interpolation. 
Inspired by the representation theory~\cite{smola2004tutorial}, complex signals in the low-dimensional space \eg, images, can be transformed as linear representations in the high-dimensional space \eg, features. From this theory, we can observe that the spatial-temporal feature $F_{ST}$ is indeed the high-dimensional feature representation for the low-dimensional image.
We use linear interpolation to obtain features at the queried coordinate $p_q$.
In the experiments of Table~\ref{tab:discussion_of_framework}, we have compared several interpolation methods,~\eg, nearest sampling, and the results show that linear interpolation shows the best performance.

In summary, for any scale $s$ and timestamp $t$, the feature interpolation (\ie, 3D to 2D) process can be formulated by Eq.~\ref{eq:st_to_sf}, given the 3D coordinate grid $C_{t,s}$ and spatial-temporal feature $F_{ST}$.
{\setlength\abovedisplayskip{3pt}
\setlength\belowdisplayskip{3pt}
\begin{equation}
    \begin{aligned}
    F_{SF} = f_{sample}(F_{ST}, C_{t,s})
    \end{aligned}
    \label{eq:st_to_sf}
\end{equation}}

\noindent \textbf{Implicit Representation Decoding:}
Finally, the sampled 2D feature map $F_{SF}$  and the temporal feature map $F_{T}$ are added together and fed into the decoder. 
For simplicity and efficiency, we design three-layers CNN structure as the decoder.
This is supported by the empirical experiment in Table~\ref{tab:discussion_of_framework}, showing that a sample CNN block can achieve good results with low complexity.

\subsection{Loss Function}
\label{subsec:loss_function}

We employ the \textit{Charbonnier loss}\cite{lai2018fast} as our VSR supervision loss $\mathcal{L}_{SR}$ between the ground truth (GT) HR frame $I_{t,s}^{HR}$ and the output SR frame $I_{t,s}^{SR}$ at timestamp $t$ with up-sampling scale $s$, as shown in Eq.~\ref{eq:frame_reconstruction_loss}, where $\epsilon$ is $1e-3$.

In training, the value range of $t$ is all timestamps of key frames $\{t_0, t_1 ... t_n\}$.
$s$ could be a real number in the range $[1.0, s_{max}]$, where $s_{max}$ is the maximum up-sampling scale during training depending on the resolution of training data.
The loss function $\mathcal{L}$ is shown in Eq.~\ref{eq:overall_loss_function}.
For example, when the resolution of input LR frame is set to be $128\times 128$ and the GT HR frame is $1024\times 1024$, the $s_{max}$ is $8$.
{\setlength\abovedisplayskip{3pt}
\setlength\belowdisplayskip{3pt}
\begin{equation}
     \begin{aligned}
        \mathcal{L}_{SR} = \sqrt{(I_{t,s}^{SR} - I_{t,s}^{HR})^2 + \epsilon ^ 2 }
    \end{aligned}
    \label{eq:frame_reconstruction_loss}
\end{equation}}

{\setlength\abovedisplayskip{-6pt}
\setlength\belowdisplayskip{3pt}
\begin{equation}
       \begin{aligned}
        \mathcal{L} = \sum_{t\in {t_0, t_1...t_n}} \left(\sum_{s\in [1.0, s_{max}]} \mathcal{L}_{SR} (I_{t,s}^{SR}, I_{t,s}^{HR})\right)
    \end{aligned}
    \label{eq:overall_loss_function}
\end{equation}}

\subsection{Real-world Dataset Collection}

Existing datasets, \eg, CED\cite{scheerlinck2019ced}, suffer from the limited resolution ($346\times 260$) and severe noise, as shown in Fig.\ref{fig:NoiseCompare}.
Although CED dataset provides the active pixel sensor (APS) frames, they are in low quality because they are simply demosaiced by OpenCV \cite{pisarevsky2010introduction} from RAW data.
Therefore, collecting HR and high-quality datasets with spatially aligned frames and events is important to inspire more research for the event-guided VSR problem.
In this paper, we collected a new real-world dataset, called ALPIX-VSR, using a ALPIX-Eiger event camera\footnote{https://www.alpsentek.com/product}.  
The camera outputs well aligned  RGB frames and events. The RGB frames enjoy a resolution of $3264\times 2448$ and are generated by a carefully designed image signal processor(ISP) from RAW data with the Quad Bayer pattern~\cite{cho2022pynet}, and the events have a resolution with $1632\times 1224$.

Our ALPIX-VSR dataset consists of 26 video sequences with 5388 frames and well-aligned events in total.
These sequences include diverse scenes, \eg, streets, buildings, flowers, textures, and machines.
To avoid motion blur and low-light noise, we collect the dataset in bright indoor and sunny outdoor scenes.
\textit{For more details about our real-world dataset, please refer to the supplementary material.}

\section{Experiments}
\subsection{Experiments Setting}
\noindent\textbf{Implementation Details and Datasets:} 
For all experiments, we use the Adam optimizer\cite{kingma2014adam} with a learning rate of $1e-4$ for CED dataset and $5e-5$ for our ALPIX-VSR dataset. We train our framework for 100 epochs with a batch size of 2 using two NVIDIA RTX A30 GPU cards.

\noindent\textbf{Evaluation Metrics:}
We statistically assess the effectiveness of our approach using the peak-signal-to-noise ratio (PSNR) and structural similarity (SSIM).

\noindent \textbf{Datasets:} We use the CED\cite{scheerlinck2019ced} and the ALPIX-VSR dataset for experiments. 
\textbf{1) CED Dataset.} 
It includes a collection of color events and video sequences in many scenes, \eg, indoor, outdoor, driving, human, calibration. The resolution of the frames and events is $346\times 260$. We follow the setting of E-VSR\cite{jing2021turning} to preprocess this dataset.  Note that the RGB frames provided by CED are obtained by demosaicing \cite{pisarevsky2010introduction} from raw frames and suffer from severe noise. 
\textbf{2) ALPIX-VSR Dataset.}
We select 20 videos for training and 6 videos for testing. The training and testing sets include 4212 and 1176 frames with aligned events, respectively.
Note that we apply data augmentation strategy, such as random crop, to ALPIX-VSR dataset for all compared methods to avoid memory overflow during training.

\begin{table*}[t!]
\renewcommand{\tabcolsep}{4pt}
\centering

\vspace{-10pt}
\resizebox{\linewidth}{!}{
\begin{tabular}{c||c c c c c c c }
\toprule
Clip Name               & DUF\cite{jo2018deep}*              & TDAN \cite{tian2020tdan}             & SOF \cite{wang2020deep}              & RBPN\cite{Haris2019rbpn}                  &  VideoINR \cite{chen2022videoinr}* & E-VSR \cite{jing2021turning} & Ours \\
\hline
\hline
people\_dynamic\_wave           & 32.02 / 0.9333    & 35.83 / 0.9540    & 33.32 / 0.9360    & 40.07 / 0.9868        & 27.47 / 0.8229    & 41.08 / 0.9891 & 38.78 / 0.9794 \\
indoors\_foosball\_2            & 30.55 / 0.9262    & 32.12 / 0.9339    & 30.86 / 0.9253    & 34.15 / 0.9739        & 26.03 / 0.7766    & 34.77 / 0.9775 & 38.68 / 0.9750 \\
simple\_wires\_2                & 30.08 / 0.9387    & 31.57 / 0.9466    & 30.12 / 0.9326    & 33.83 / 0.9739        & 26.77 / 0.8321    & 34.44 / 0.9773 & 38.67 / 0.9815 \\
people\_dynamic\_dancing        & 31.64 / 0.9369    & 35.73 / 0.9566    & 32.93 / 0.9388    & 39.56 / 0.9869        & 27.36 / 0.8202    & 40.49 / 0.9891 & 39.06 / 0.9798 \\
people\_dynamic\_jumping        & 31.57 / 0.9334    & 35.42 / 0.9536    & 32.79 / 0.9347    & 39.44 / 0.9859        & 27.24 / 0.8183    & 40.32 / 0.9880 & 38.93 / 0.9792 \\
simple\_fruit\_fast             & 37.46 / 0.9442    & 37.75 / 0.9440    & 37.22 / 0.9390    & 40.33 / 0.9782        & 27.21 / 0.8456    & 40.80 / 0.9801 & 41.96 / 0.9821 \\
outdoor\_jumping\_infrared\_2   & 25.33 / 0.8162    & 28.91 / 0.9062    & 26.67 / 0.8746    & 30.36 / 0.9648        & 26.88 / 0.8226    & 30.70 / 0.9698 & 38.03 / 0.9755 \\
simple\_carpet\_fast            & 31.43 / 0.8811    & 32.54 / 0.9006    & 31.83 / 0.8774    & 34.91 / 0.9502        & 24.21 / 0.5909    & 35.16 / 0.9536 & 36.14 / 0.9635 \\
people\_dynamic\_armroll        & 31.38 / 0.9311    & 35.55 / 0.9541    & 32.79 / 0.9345    & 40.05 / 0.9878        & 27.26 / 0.8193    & 41.00 / 0.9898 & 38.84 / 0.9787 \\
indoors\_kitchen\_2             & 29.92 / 0.9273    & 30.67 / 0.9323    & 29.61 / 0.9192    & 31.51 / 0.9551        & 26.44 / 0.7502    & 31.79 / 0.9586 & 37.68 / 0.9726 \\
people\_dynamic\_sitting        & 30.62 / 0.9331    & 35.09 / 0.9561    & 32.13 / 0.9367    & 39.03 / 0.9862        & 27.63 / 0.8230    & 39.97 / 0.9884 & 38.86 / 0.9810 \\
\hline
\hline
average PSNR/SSIM               & 31.09 / 0.9183    & 33.74 / 0.9398    & 31.84 / 0.9226    & 36.66 / 0.9754    &26.77 / 0.7938 & 37.32 / \textbf{0.9783} & \textbf{38.69} / 0.9771 \\
\bottomrule
\end{tabular}
}
\vspace{-8pt}
\caption{
Quantitative results (PSNR/SSIM) of the proposed our framework and other methods on the CED dataset for $\times 2$.
Because the official training code is not available, * denoted values were acquired from the pre-trained model that the authors have released.}
\label{tab:framework_comparison_ced_2x}
\end{table*}

\begin{figure*}[t!]
\centering
\includegraphics[width=1\linewidth]{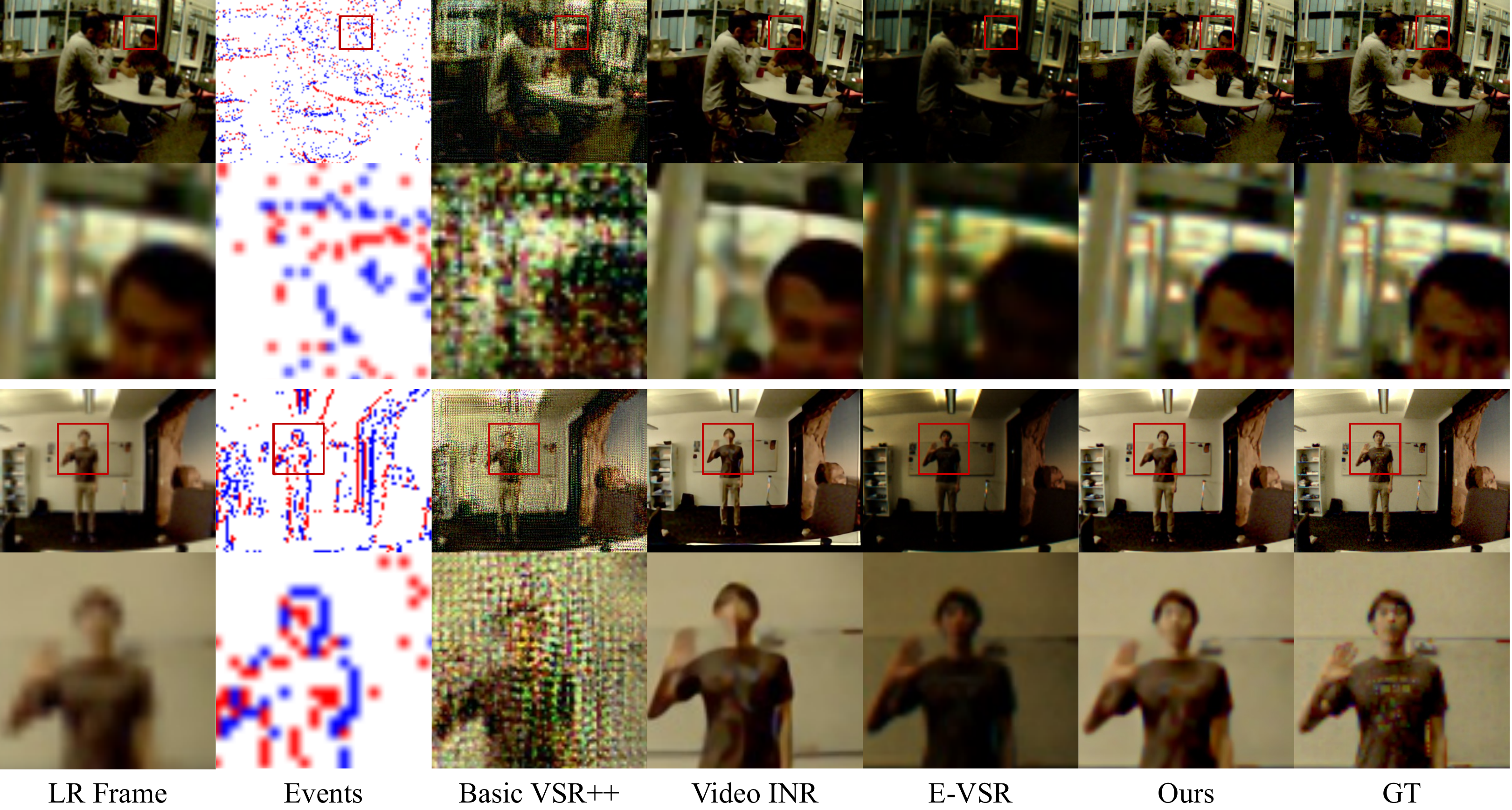}
\vspace{-10pt}
\caption{Visual results of $\times 4$ VSR on the CED dataset. Our method recovers more details, \eg, textual, edge than the SoTA event-guided VSR method E-VSR\cite{jing2021turning} and the random up-sampling VSR method Video INR\cite{chen2022videoinr}.}
\label{fig:fig4}
\centering
\vspace{-8pt}
\end{figure*}

\subsection{Comparison with SoTA Methods}

We compare our method with six SoTA methods under three VSR settings: 
(I) one SoTA event-guided, fixed-scale VSR method E-VSR\cite{jing2021turning}
(II) one SoTA method of frame-based arbitrary-scale VSR method VideoINR \cite{chen2022videoinr}
(III) five SoTA methods of frame-based, fixed-scale VSR methods: BasicVSR++ \cite{chan2021basicvsr++}, RBPN \cite{Haris2019rbpn}, SOF \cite{wang2020deep}, TDAN\cite{tian2020tdan}, DUF\cite{jo2018deep}.
We report $\times 2$ and $\times 4$ super-resolution results of our method and all 6 comparison methods on CED dataset. We also compare our method with E-VSR and BasicVSR++ on our ALPIX-VSR dataset. Moreover, we compare our method with VideoINR on out-of-distribution scales to demonstrate our method's ability for arbitrary-scale VSR.
Note that E-VSR only supports $\times 2 $ and $\times 4$ SR, and BasicVSR++ only supports $\times 4 $ SR.

\noindent\textbf{Evaluation on CED Dataset}
Table~\ref{tab:framework_comparison_ced_2x} and Table~\ref{tab:framework_comparison_ced_4x} present the quantitative results for $\times 2$ and $\times 4$ VSR, respectively. Our model clearly outperforms other methods in terms of PSNR, and shows a comparable performance in SSIM with E-VSR. The qualitative results in Fig. \ref{fig:fig4} demonstrate that our model is capable of recovering fine details, like sharp edges and detailed textures. 
We can see that event-guided methods (E-VSR and our method) yield better performances than the frame-based counterparts, showing the complementary effects of event data for VSR. 
Furthermore, VideoINR performs noticeably worse than other methods, indicating that frame-based implicit neural representations struggle with the low-resolution and high-noise input.
Contrarily, our approach benefits from additional event information and generates satisfactory INRs.

Table~\ref{tab:framework_comparison_ced_4x} shows quantitative results in $\times 4$ SR on the CED dataset, where our model achieves SoTA performance while remaining lightweight and efficient. 
Specifically, our model only accounts for one-200th of the E-VSR. 
BasicVSR++, the SoTA method of frame-based VSR, fails to perform well on CED dataset as the upsampling scale increases, which demonstrates its less robustness on the highly noisy dataset.

\begin{table}[t!]
\centering
\footnotesize
\vspace{-1pt}
\setlength{\tabcolsep}{0.045\linewidth}{
\begin{tabular}{c|c c c}
\hline
Methods                                 & Model Size($M$)    & PSNR  & SSIM \\
\hline
DUF \cite{jo2018deep}                   & 1.90          & 24.43 & 0.8177    \\
TDAN \cite{tian2020tdan}                & 1.97          & 27.88 & 0.8231    \\
SOF \cite{wang2020deep}                 & 1.00          & 27.00 & 0.8050    \\
RBPN \cite{Haris2019rbpn}               & 12.18         & 29.80 & 0.8975    \\
BasicVSR++ \cite{chan2021basicvsr++}    & 7.30          & 14.76 & 0.1641     \\
VideoINR \cite{chen2022videoinr}*       & 11.31         & 25.53 & 0.7871   \\
E-VSR  \cite{jing2021turning}           & 412.42        & 30.15 & 0.9052    \\ 
Ours                                    & 2.45          & \textbf{31.12} & \textbf{0.9211}    \\
\hline
\end{tabular}
}
\vspace{-8pt}
\caption{\textbf{Quantitative results on CED dataset} for $\times 4$. * denotes the values obtained from the official pre-trained models.}
\label{tab:framework_comparison_ced_4x}
\end{table}

\begin{figure*}[t!]
\centering
\includegraphics[width=0.96\linewidth]{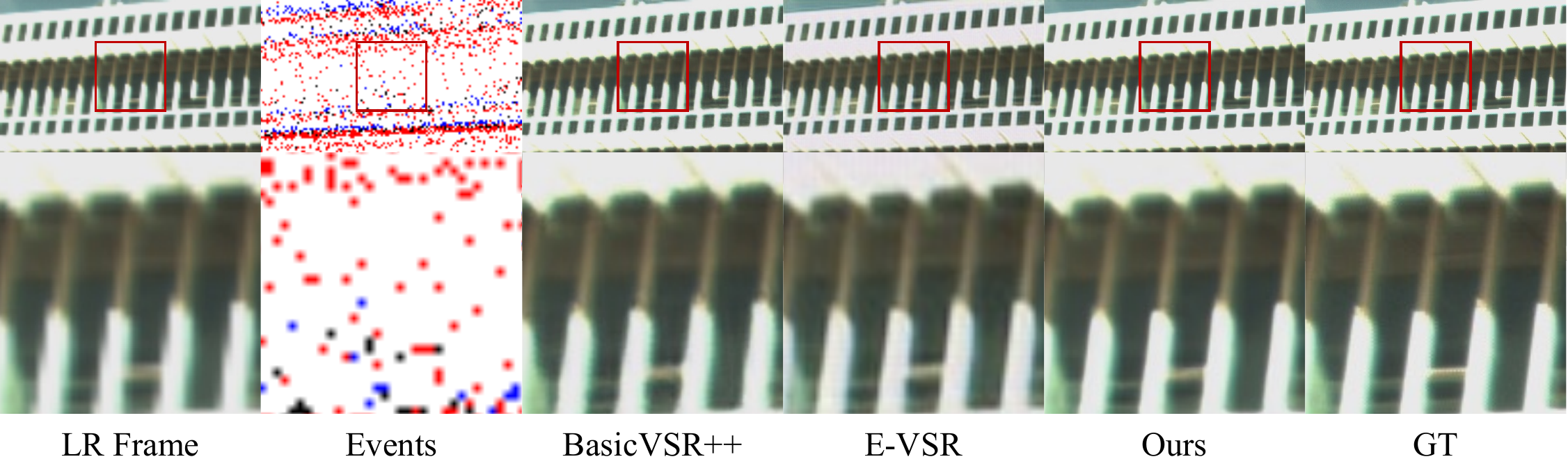}
\vspace{-10pt}
\caption{Results of $\times 4$ VSR on the ALPIX-VSR dataset.
}
\label{fig:APLIX-4x-v2k}
\centering
\vspace{-10pt}
\end{figure*}

\begin{figure*}[t!]
\centering
\includegraphics[width=0.97\linewidth]{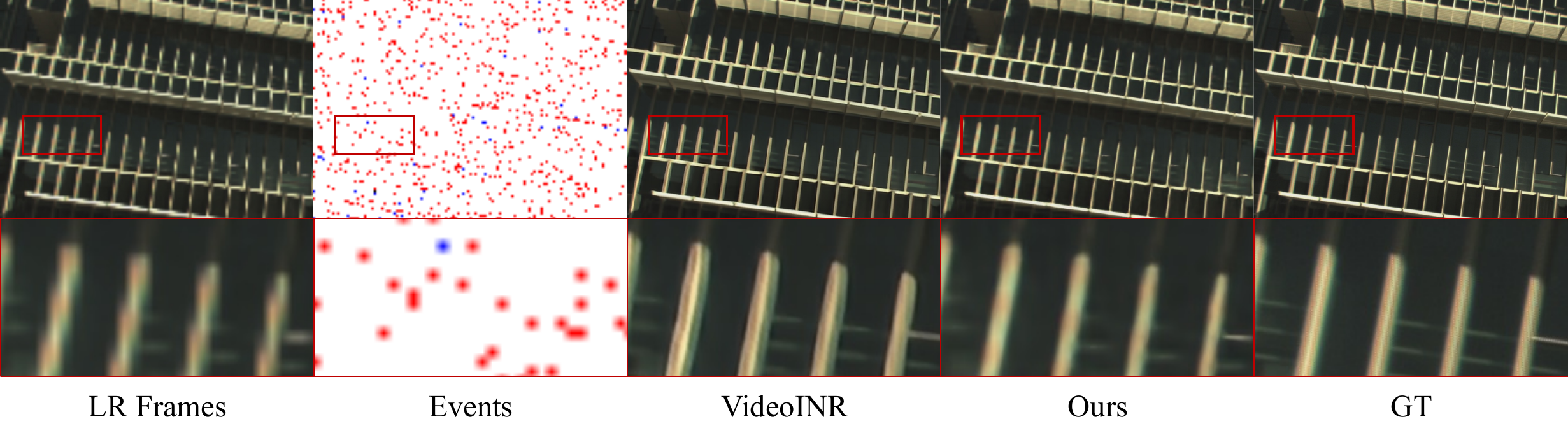}
\vspace{-10pt}
\caption{Results of $\times 8$ VSR on the ALPIX-VSR dataset.}
\label{fig:aplix_results}
\centering
\vspace{-8pt}
\end{figure*}

\noindent\textbf{Evaluation on the ALPIX-VSR Dataset}
The quantitative and qualitative results are shown in Table~\ref{tab:alpix-2458} and Fig.~\ref{fig:aplix_results}.
We present comparison results of our method with BasicVSR++ and E-VSR on our collected real-world dataset.

\begin{table}[t!]
 \footnotesize
 \centering
\vspace{-10pt}
\setlength{\tabcolsep}{0.07\linewidth}{
\begin{tabular}{c c|c c c}
\hline
                & Methods       & PSNR              & SSIM \\
\hline
                &E-VSR          &36.10              & 0.9761       \\
$\times 2$      %&VideoINR       &29.22              & 0.9010 \\                
                &Ours           &\textbf{38.25}      & \textbf{0.9822}      \\
\hline
                & E-VSR         & 32.54             & 0.9163       \\
$\times 4$      & BasicVSR++    & 35.30             & 0.9353  \\
                % & VideoINR*       & \textbf{37.21}     & \textbf{0.9539}    \\
                & Ours          & \textbf{37.12}     & \textbf{0.9503}      \\
\hline
\hline
$\times 6$      & VideoINR*       & 31.15                & 0.9084 \\
                & Ours           & \textbf{31.85}       & \textbf{0.9267}      \\
\hline          
$\times 8$      &VideoINR*       &  28.11  &  0.8625  \\     
                & Ours          & \textbf{28.53}      & \textbf{0.8901}      \\
\hline
\end{tabular}
}
\vspace{-5pt}
\caption{\textbf{Quantitative comparison} (PSNR/SSIM) of our methods and other methods on the ALPIX-VSR dataset. * denotes the values obtained from the official pre-trained models.}
\vspace{-10pt}
\label{tab:alpix-2458}
\end{table}

\vspace{-5pt}
\subsection{Random Scale Up-sampling}
\vspace{-5pt}
Results of random scale upsampling are shown in Table~\ref{tab:alpix-2458}.
We upsample the video frames to $\times 2$ , $\times 4$, $\times 6$  , $\times 8$, and compare with the SoTA models, E-VSR and BasicVSR++. 
Though these two models are strictly constrained to a specific upsampling scale, our method presents superior performance on these settings. 

We also compare our method with VideoINR, the SoTA random-scale VSR method quantitatively in $\times 6$, $\times 8$. 
Results in Table \ref{tab:alpix-2458} show that our model surpasses VideoINR in all evaluated scales of $\times 6$, $\times 8$. 
Furthermore, to evaluate the performance of our method on arbitrary scales, we conduct experiments of 6 random-chosen float scales.
The results are shown in Table~\ref{tab:table5} and indicate our model's robustness across arbitrary random scales.
It is easy to find that values of performance, \eg, PSNR, SSIM, and upsampling scales are not strictly monotonic.

\begin{table}[t!]
  \centering
  \footnotesize
\setlength{\tabcolsep}{0.03\linewidth}{
\begin{tabular}{c|ccc}
\hline
                & $\times 1.8$        & $\times 2.6$      & $\times 5.6$      \\
\hline
Ours            & 39.2508 / 0.9803    & 37.3408 / 0.9589    & 31.2549 / 0.9135              \\

\hline
\hline
                & $\times 6.6$      & $\times 7.1$      & $\times 7.8$ \\
\hline
Ours            & 28.3182 / 0.8772  & 28.3188 / 0.87762  & 28.3198 / 0.8783    \\
\hline
\end{tabular}
}
\vspace{-5pt}
\caption{\textbf{Quantitative results}(PSNR/SSIM) of random-scale comparison on the\textbf{ALPIX-VSR dataset}.}
\label{tab:table5}
\vspace{-15pt}
\end{table}

\subsection{Ablation Studies and Discussion}\
\label{ablation_discussion}
\vspace{-2pt}
The following ablation experiments investigate the importance of each of our proposed modules.
As it takes more than $120$ hours to train a model on the complete CED dataset, we uniformly select $1/5$ of CED as the dataset for ablation experiments. 

\begin{figure}[t!]
\centering
\includegraphics[width=\linewidth]{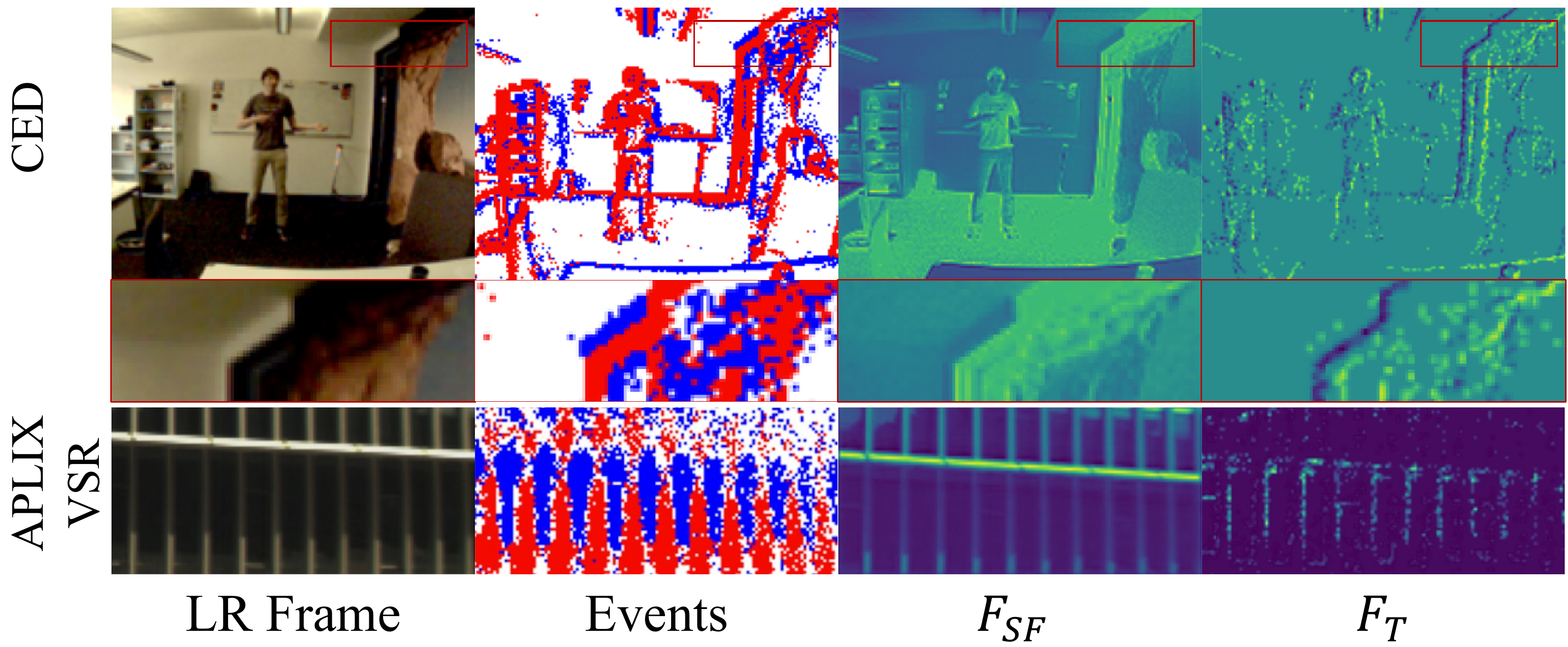}
\vspace{-20pt}
\caption{\textbf{Feature visualization} of $F_{SF}$ (Eq.~\ref{eq:st_to_sf}) and $F_{T}$ (Eq.~\ref{eq:tf}).}
\label{fig:TwoBranch}
\centering
\vspace{-5pt}
\end{figure}

\noindent \textbf{Efficiency of TF branch and Shallow Feature Fusion:}
Table~\ref{tab:ablation_TF_branch_shallow_feature_fusion} validates the contribution of TF branch and shallow feature fusion in STF branch.
The removal of TF branch reduces both PSNR and SSIM scores, where PSNR drops by $0.18dB$. In Fig.~\ref{fig:TwoBranch}, we use PCA~\cite{daffertshofer2004pca} to visualize the output of TF branch. 
As can be seen, $F_{T}$ focuses on edge and corner information, which is very helpful for VSR, especially texture recovery.
We also find that removing shallow feature fusion results in a large performance drop (PSNR drops by 0.38 $dB$ and SSIM drops by nearly 0.1 $dB$). This finding indicates that fusion on shallow features is critical since shallow features carry rich local structure information which may be missing in deep features.

\begin{table}[t!]
  \centering
  \footnotesize
\resizebox{\linewidth}{!}{
\begin{tabular}{c|ccc|cc}
\hline
    & \makecell[c]{TF\\ Branch}         & \makecell[c]{Shallow\\ Feature Fusion} & \makecell[c]{Model\\ Size($M$)}     & PSNR  & SSIM   \\
\hline
\hline
1    & w                & w                      & 2.4513  & 38.14 & 0.9820 \\
\hline
2    & \textbf{w/o}     & w                      & 2.4482  & 37.96 & 0.9812 \\
3    & w                & \textbf{w/o}           & 1.7360  & 37.76 & 0.9729 \\
\hline
\end{tabular}
}
\vspace{-8pt}
\caption{
Ablation of the TF branch and shallow feature fusion.
}
\vspace{-5pt}
\label{tab:ablation_TF_branch_shallow_feature_fusion}
\end{table}

\begin{table}[t!]
  \centering
\footnotesize

\resizebox{\linewidth}{!}{
\begin{tabular}{c|ccc|cc}
\hline
     & Interpolation        & Decoder                   & LF Channels   & PSNR  & SSIM   \\
\hline
\hline
1    & Linear                & CNN                      & 16            & 38.14 & 0.9820 \\
\hline
2    & \textbf{Nearest}      & CNN                      & 16            & 28.91 & 0.9131 \\
3    & Linear                & \textbf{SIREN}           & 16            & 10.30 & 0.2997 \\
4    & Linear                & \textbf{MLP}             & 16            & 37.94 & 0.9811 \\
5    & Linear                & CNN                      & \textbf{8}    & 36.82 & 0.9763 \\
6    & Linear                & CNN                      & \textbf{24}   & 38.25 & 0.9825 \\
\hline
\end{tabular}
}
\vspace{-8pt}
\caption{\textbf{Impacts of interpolation methods, decoder designs and channel size} of the spatial-temporal features $F_{ST}$.}
\label{tab:discussion_of_framework}
\vspace{-6pt}
\end{table}

\noindent \textbf{Feature Interpolation:}
We apply a feature-based interpolation strategy, \ie interpolating features near a coordinate and sending the interpolated feature into the decoder to reconstruct HR frames. 
Such strategy has been studied in a prior work on implicit neural representation learning for 3D objects \cite{sun2022direct} and shown to be able to recover clearer details and sharper edges.
We further study the influence of interpolation methods, as shown in Table~\ref{tab:discussion_of_framework}. The trilinear manner of interpolation yields better PSNR and SSIM scores compared with the nearest interpolation.  

\noindent \textbf{Feature Decoder:}
We also compare the performance of different decoder designs, including MLP, SIREN\cite{sitzmann2020siren} and CNN. Table~\ref{tab:discussion_of_framework} shows that decoding with CNN has the best performance among these methods, while non-convergence occurs with SIREN.
We argue the reason CNN performs well is that decoding in STIR is only a dimensional reduction process, so no complicated design is required.%, \eg, SIREN.

\begin{figure}
\centering
\includegraphics[width=0.9\linewidth]{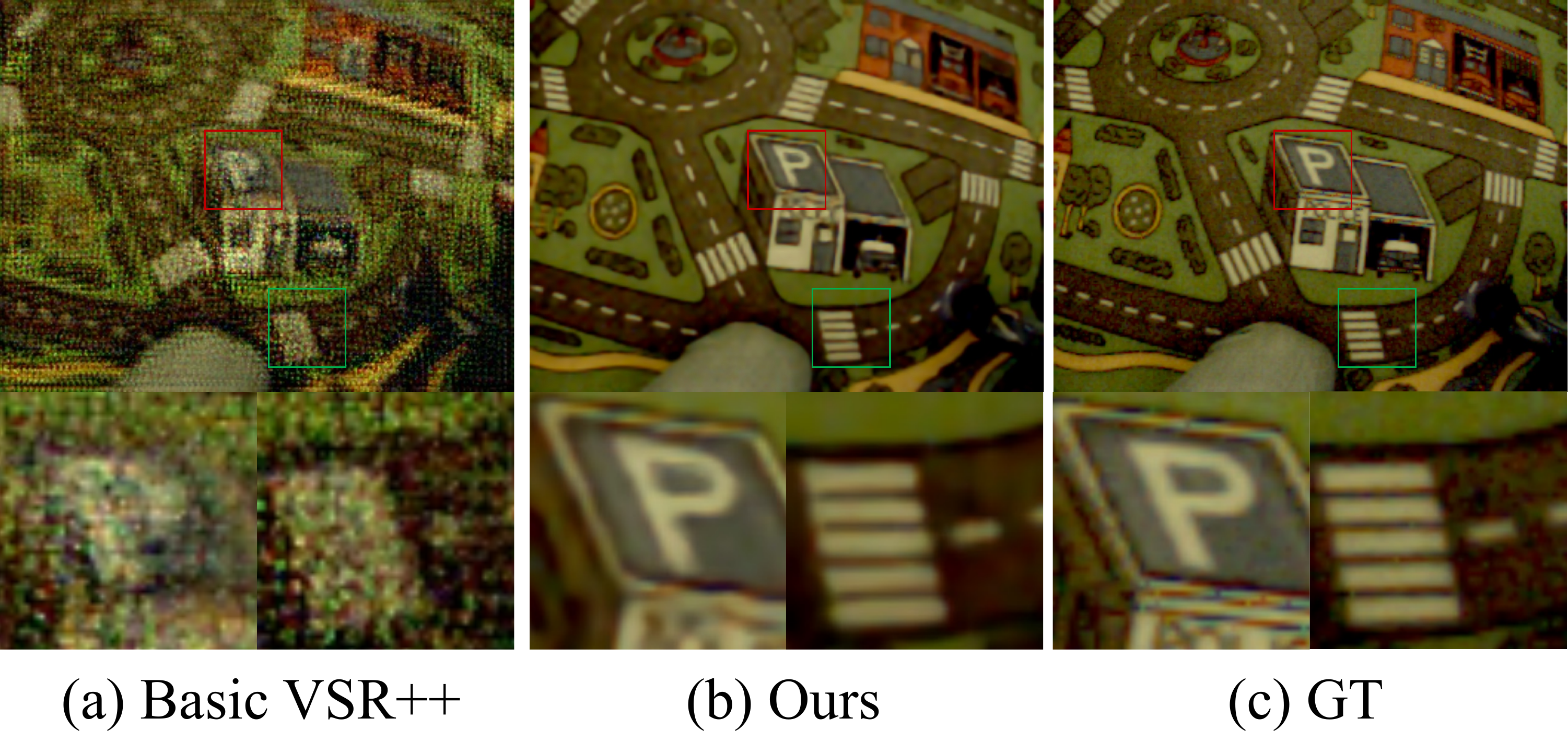}
\vspace{-5pt}
\caption{\textbf{Comparison of noise removal capacity} of BasicVSR++\cite{chan2021basicvsr++} with our methods, with respect to  HR GT.}
\label{fig:NoiseCompare}
\centering
\vspace{-10pt}
\end{figure}

\noindent \textbf{Robustness to Noise:}
In comparison to BasicVSR++, our method not only performs SR but also removes noise on the CED dataset, as shown in Fig.\ref{fig:NoiseCompare}.
Note that BasicVSR++ is a SoTA frame-based method in VSR task.
We analyze that the poor performance of BasicVSR++ on CED is caused by the dependence for only frame modality and excessive emphasis on the frame's high-frequency information.
High frequencies are often present in the image as edges, corners and noise.
Therefore BasicVSR++ is easily affected by serious noise. In comparison to BasicVSR++, our framework has more robustness.
Benefiting from the guidance of events, \eg, edges, corners, our method can effectively reduce the adverse effects of noise on frames.

\begin{table}[t!]
\setlength{\tabcolsep}{10mm}
 \centering
\footnotesize
\begin{tabular}{c|cc}
\hline
                        & PSNR      & SSIM \\ 
\hline
3 to 1                  & 38.14     & 0.9820     \\
5 to 3                  & 38.04     & 0.9818  \\
\hline
\end{tabular}
\vspace{-5pt}
\caption{\textbf{Ablation for the number of input and output frames}.}
\vspace{-5pt}
\end{table}

\section{Conclusion}
In this paper, we proposed a novel framework which jointly learns INRs from RGB frames and events, and enables arbitrary scale VSR. 
Our method effectively uses high temporal resolution property of events to complement RGB frames with STF and TF. 
A simple yet effective STIR is used to recover frames at arbitrary scales.
Extensive experiments on two real-world datasets validate our method enjoys better performance over current related SoTA methods with significantly lower model size.

% \noindent \textbf{Limitation and Future Work:} In this work we tackle the problem of spatial VSR. Since our work learns INRs of videos, it naturally enables frame interpolation, 
% which will be explored in our future work. We hope our work can shed light on more research about enhancing implicit neural representations with non-photorealistic data.

\section{Acknowledgment}
This work was supported by the Research Project Fund of AlpsenTek and the National Natural Science Foundation of China (NSFC) under Grant No. NSFC22FYT45. 

\clearpage
{\small
\bibliographystyle{ieee_fullname}
\bibliography{cvpr23}
}

\end{document}